\documentclass[11pt]{article}

\usepackage[left=0.98in, right=0.98in, top=1.0in, bottom=1.0in]{geometry}

\newcommand\beq{\begin{equation}}
\newcommand\eeq{\end{equation}}

\newcommand\bmat{\begin{bmatrix}}
\newcommand\emat{\end{bmatrix}}

\usepackage{hyperref}       % hyperlinks
\usepackage{url}            % simple URL typesetting
\usepackage{booktabs}       % professional-quality tables
\usepackage{amsfonts}       % blackboard math symbols
\usepackage{subcaption}

\usepackage{graphicx}
\usepackage{wrapfig}
\usepackage{cite}   % compact citations to multiple papers eg [2-5]
\usepackage{amsmath}

\usepackage{caption}   % better spacing of Table 1
\newcommand{\Sel}{\textit{Selector~}}
\newcommand{\Seq}{\textit{Sequence~}}
\newcommand{\Par}{\textit{Parallel~}}
\newcommand{\Rus}{\textit{Repeat until Success~}}
\newcommand{\Dec}{\textit{Decorator~}}
\newcommand{\Suc}{\textit{Success~}}
\newcommand{\Fail}{\textit{Failure~}}

\title{Hidden Markov Models derived from Behavior Trees}

\author{%
  Blake Hannaford\\%\thanks{Use footnote for providing further information
  Department of Electrical \& Computer Engineering\\
  University of Washington\\
  Seattle, WA 98195 \\
  \texttt{blake@uw.edu} \\
}

\begin{document}

\maketitle
% 
%   The abstract paragraph should be indented \nicefrac{1}{2}~inch (3~picas) on
%   both the left- and right-hand margins. Use 10~point type, with a vertical
%   spacing (leading) of 11~points.  The word \textbf{Abstract} must be centered,
%   bold, and in point size 12. Two line spaces precede the abstract. The abstract
%   must be limited to one paragraph.
%   
\begin{abstract}
  Behavior trees are rapidly attracting interest in robotics
  and human task-related motion tracking.  
  However no algorithms currently exist to track or identify parameters 
  of BTs under noisy observations.  We report a new relationship between BTs, augmented
  with statistical information, and Hidden Markov Models.  Exploiting this relationship 
  will allow application of many algorithms for HMMs (and dynamic Bayesian networks)
  to data acquired from BT-based systems. 
  
  {\bf Submitted to IEEE Transactions on Robotics and Automation, 23-Jul-2019.}
\end{abstract}

%%%%** Section 1 
\section{Introduction}

The domain of this paper is tools for representing, observing, and tracking task 
performance in which 
the observations or measurements of the task progression are noisy.  Applications 
include monitoring progression of defined tasks performed by a human through noisy measurements. 
Hidden Markov Models (HMMs) are widely used and well known in applications such as 
speech\cite{juang1985mixture,juang1991hidden}, surgery\cite{reiley2011review}, and 
bioinformatics\cite{baldi1994hidden}.
Behavior Trees (BTs) are a plan representation, 
arising in the world of Game AI for non-player characters, which are making inroads in robotics
\cite{paxton2017costar,Colledanchise2017Modularize} and medicine\cite{BRLM019}.  

The contribution of this paper is a new relationship between BTs and HMMs.  
We show that when a BT is
augmented with statistical or prior probability information, then it corresponds to a unique 
HMM having a specific structure.  
We can then exploit this link and the well known HMM algorithms for state estimation and 
system identification to monitor the evolution of a system described by a BT in
spite of noisy observations.
Useful links then also follow between BTs and Dynamic Bayesian Nets which could improve 
the power of these methods\cite{ghahramani2001introduction}.

%%%%** Section 1.1
\subsection{Models}\label{ModelReview}
% 
% 

% % HMMs - sparse, manually initialized, observations discretized with VQ.
% In this work, a HMM (Figure \ref{basichmm}) is can be described
% \[
% HMM == \{ A_{n\times n}, B_{n\times m} \}
% \]
% where $n$ is the number of HMM states and $m$ is the number of observation symbols.

In this work, a HMM having $N$ states and $J$ discrete output symbols is described by
\[
HMM = \left \{
\begin{array}{ll}
\Pi_i &\in \mathcal{R}^N\\
A_{i, j} & \in \mathcal{R}^{N\times N} \\
B_{i,j} & \in \mathcal{R}^{N\times J}
\end{array}
\right .
\]

Where $\Pi_i$ is the probability of the initial state being state $i$, 
$A_{i,j}$ is the probability of transition from state $i$, to state $j$,
and $B_{i,j}$ is the probability of symbol $j$ being observed in state $i$.
 The observations (also called emissions) depend probabilistically on the current state
(which is thus not directly observable). 

Early work on Hidden Markov Models made a major early advance in speech  recognition\cite{juang1985mixture,juang1991hidden}. 
Three well-known algorithms,  reviewed in\cite{rabiner1989tutorial} and widely used with HMMs,
provide useful information from signals arising from a task performance. 
The Baum-Welch expectation-maximization algorithm tunes the parameters of a 
HMM to increase the probability that a 
given data set arises from that HMM. 
The Viterbi algorithm optimally estimates the most likely state   sequence from a 
given data set based on a HMM,
and the Forward Algorithm computes the probability that a given data set was produced by a given HMM. In robotics,
Hidden Markov Models have been applied to modeling sequential data arising from manipulation processes\cite{BRL042,BRL052, 
yang1994hidden, hovland1996skill,reiley2011review,paxton2017costar}.
% Despite rapidly
% growing interest in Recurrent Neural Nets (RNN) for classification of time series data,
% \cite{panzner2016comparing} recently compared HMMs to RNNs in an activity recognition task and found 
% faster parameter learning 
% for the HMM, when training data was limited, with equivalent classification performance.   
 
\paragraph{Vector Quantization (VQ)}
When the random observations 
are high dimensional (such as six-axis or more motion/force/torque readings measured experimentally in surgery\cite{BRL199}), they may be compressed into discrete
symbols using a vector-quantizer such as K-means clustering\cite{rabiner1989tutorial, BRL174}.
\cite{BRL174} found
that a code-book size of 63 code-words performed well with 9-dimensional sensor data recorded from 
a bowel suturing surgical task.

%%%%%%%%%%%%%%%%%%%%%%%%%%%%%%%%%%%%%%%%%%%%%%  BTs

%%%%** Section 2 
\section{BTs and augmented BTs}

\paragraph{BT background}
Prior to about 2010, the term ``Behavior Tree'' was used idiosyncratically by several authors, but around that time a body of literature began to emerge around a tree model of behaviors used by the video game industry for AI-based non-player characters\cite{halo,lim2010evolving}.  These BTs assume that units of intelligent behavior (such as decisions or units of action) can be described such that they perform a piece of an overall task/behavior, and that they can determine and return a 1-bit result indicating success or failure.  These units are the leaves of BTs.
The level of abstraction of BT leaves is not specified by the BT formalism and may vary from one application to another or within a single BT.   
But BTs are deterministic and do not have well established tools for tracking with 
noisy data, or parameter identification.
% 
% In the context of
% medicine\cite{BRL_Bts_in_Med}, 
% BT leaves can be diagnostic or therapeutic steps such as the administration of a blood test or a
% step of a surgical procedure.   In describing patient management that might occur over several days, a BT leaf
% might describe a sub-procedure such as to perform a biopsy, but that biopsy could in-turn be broken down into its
% own BT.

%  key intro material from Mark Whipple
% Specific detailed medical procedures are often published 
% (by for example practitioners, institutions, or professional
% society working groups) in terms of ``clinical practice guidelines".  
% Over the past two decades a number of computer-interpretable guideline (CIG) formalisms 
% have been developed
% in the Medical Informatics literature, including rule-based (Arden syntax), 
% logic-based (PROforma), network-based (PRODIGY), and workflow-based (GUIDE) models (for a detailed 
% review see\cite{OpenClinical}).   Compared to these models, 
% the ABTs we develop below are simpler, and more clearly applicable to theoretical analysis.  
% Many CIG implementation have instead focused on 
% interoperability  with online archives, electronic medical record systems, 
% and clinical decision support systems.  Our focus here instead is on formal links between, as one example,
% a BT-based CIG, and stochastically corrupted data, to study and monitor medical procedures.

% 
In medical robotics, researchers are turning attention to augmentation of the purely teleoperated  existing
systems such as the daVinci$^{TM}$ surgical robotic system (Intuitive Surgical, Sunnyvale, CA) with intelligent
functions\cite{BRL278,BRL279}.  In this context, BT leaves can represent 
autonomous robotic functions such as a guarded move,
a precision cutting action, acquisition of an ultrasound image, creation of a plan, etc.
Earlier medical robotics systems containing automation, such as Robodoc\cite{kazanzides1992}
represented task sequences with scripting languages.

Recent literature has applied BTs to UAV control\cite{Ogren2012}, 
humanoid robotic control\cite{tumova2014maximally}, and human-robot cooperation 
in manufacturing\cite{paxton2017costar}. 
Theoretical classifications of BTs have 
been conducted by several authors which  formally 
relate BTs to Finite State Machines (FSMs)\cite{Colledanchise2017Modularize} yielding advantages of modularity 
and scalability with respect to FSMs.  
Other theoretical 
studies have related BTs to Hybrid Dynamical 
Systems\cite{colledanchise2014behavior},  and have developed means to guarantee 
correctness and composability of BTs\cite{colledanchise2017synthesis}. 
Software packages and 
Robotic Operating System (ROS, \cite{quigley2009ros}) 
implementations\footnote{\url{https://github.com/miccol/ROS-Behavior-Tree}} are 
now available\cite{marzinotto2014towards}. 
Several of the above authors have developed graphical 
user interfaces with which humans 
(such as domain experts) can create, visualize, and edit BTs.
The reader is referred to the above references for 
ample introductory material and examples of BT concepts.

%%%%** Figure 1 
% \begin{wrapfigure}[20]{r}{.35\textwidth}
\begin{figure}
    \centering%\vspace{-4mm}
    \includegraphics[width=0.3\linewidth]{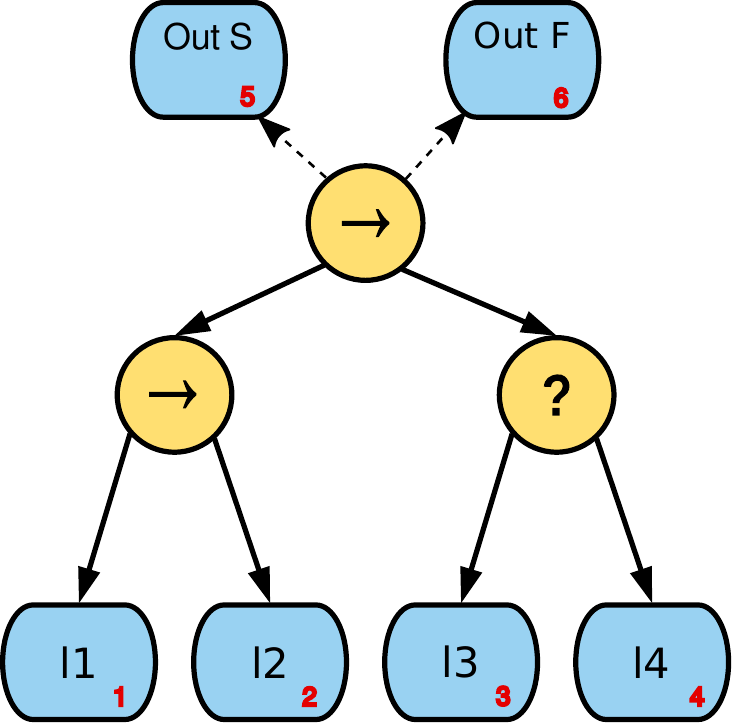}
     {\small    \caption{A simple Behavior Tree example. Leaves $l_1$ and $l_2$ belong to a \Seq ($\to$) node and 
$l_3,l_4$ to a selector (?).  If $l_1$ fails, control passes to $Out_F$. If $l_1$ and then $l_2$
succeed, control passes to the \Sel (?) node which starts $l3$ etc.}}\label{basicBT}
\end{figure}
% \end{wrapfigure}
% 
% %%%%** Figure 1 
% \begin{figure}[h]\centering
% \includegraphics[width=0.25\textwidth]{simp4stateABT.pdf}
% \caption{A simple Behavior Tree example. Leaves $l_1$ and $l_2$ belong to a \Seq node and 
% $l_3,l_4$ to a selector.  If $l_1$ fails, control passes to $Out_F$. If $l_1$ and then $l_2$
% succeed, control passes to the \Sel (?).}\label{basicBT}
% \end{figure}

When representing a process with BTs, the analyst breaks the task down into 
modules which are the  leaf nodes of the BT. Every BT node must return either 
\Suc or \Fail when called by its parent node.  All higher level nodes in the BT 
define composition rules to combine the leaves including: \Seq, \Sel, and \Par 
node types. A \Seq node defines the order of execution of leaves and returns 
\Suc if all leaves succeed in order, returning \Fail at the first child failure. 
A \Sel node (also called ``Priority" node by some authors) tries leaf behaviors 
in a fixed order, returns success  when a node succeeds,  and returns failure if 
all leaves fail. A \Par node starts all its child nodes concurrently and returns 
success if a specified fraction of its children return success. Further refining 
the behavior tree, ``\Dec'' nodes have a single child and can modify 
behavior of subsequent branches with rules such as ``repeat until Success".

Applied to robotics, BTs have been explored in the context of humanoid
robot control
\cite{marzinotto2014towards,colledanchise2014,bagnell2012integrated},
human-robot collaborative control\cite{paxton2017costar}, 
and as a modeling language for 
intelligent robotic surgical procedures \cite{BRL278,BRL279}.

%%%%** Section 2.1
\subsection{ABT definitions}\label{ABTdefs}

\paragraph{Augmented BTs}
With experience   comparing real task execution traces with a BT representing the task,
we can learn that some leaves may succeed infrequently or all the time.  
Let $iS$ be the event where leaf $i$ succeeds and $iF$ the event where it fails.
To incorporate information about $iS$ statistics, we define
an Augmented Behavior Tree, or ABT.

\beq
ABT == \{ BT, P_{l}, B_{(l+2)\times m} \}
\eeq
where $P_{l} = [ps_1, ps_2, \dots, ps_l]^T$, $ps_i = P(iS |state  = i)$ and
$l$ is the number of BT leaves or states. 
We add two   ``output states'', $O_S, O_F$ to the states defined by the BT leaves  to 
indicate the final output of the BT root node giving $l+2$ total states. 
$B_{l+1}, B_{l+2}$ are the observation probability distributions for each special output state.  
Note that these states are not nodes of the BT.
We initially assume that the observation is independent of  \Suc~and \Fail~. We can 
get initial values for $P_{l}, B_{l+1}, B_{l+2}$ from experience, or estimate 
them from expert knowledge.

For example, assuming expert-labeled sensor data is collected from experiments 
or experience, we can accumulate a frequentist estimate of
$p_i$ for each leaf by computing $\hat{p}s_i = n_{iS} / n_i $,
where $n_{iS}$ is the number of times node $i$ returned \Suc~, and $n_i$ is the 
total number of times node $i$ was executed.
% $P_l$ by computing $\hat{P}_j = N_{jS} / N_j $
% where $N_{jS}$ is the number of times node $j$ returned \Suc~, and $N_j$ is the 
% total number of times node $j$ was executed.

%%%%** Section 3 
\section{Mapping Augmented BTs to HMM}

Here we derive a few results on which we can ground the link between ABTs and HMMs. 
We restrict the types of permitted BT nodes to leaves, \Sel, and \Seq nodes.   
\Dec and \Par nodes will be considered in below (Section \ref{btrefinement})
  
% \input{raw_theory_matl.tex}
% Raw Theoretical BT Material

%%%%** Section 3.0.1 
\subsection{For each ABT, there is a unique HMM}\label{UniqueHMM}
\paragraph{Proof:}
        \begin{enumerate}
            \item Assume the ABT has $l$ leaves, numbered from left to right. Left to right is   the
            conventional  direction used by selector ($\to$) and sequence (?) nodes in the BT literature.
            
            \item Create a HMM state for each ABT leaf, and two additional terminal HMM states,
            $O_S$ and $O_F$, representing overall success or failure of the ABT, 
            so that the HMM dimension is $N = l+2$.
            
            \item By definition of BT nodes, for each leaf, $j$, belonging to a specific BT, 
            there is a unique successor leaf to which the BT will transition on success, $S_s$
            and a second leaf/HMM state to which it will transition on failure, $S_f$.
            
            \item Because of the left to right structure of Step 1, if the system is in state $j$ 
            (i.e. running leaf $j$), then
            \[
            S_s > j \; \mathrm{and} \; S_f > j
            \]
            
            \item Thus the resulting HMM transition matrix, $A$  will be $N\times N$,
            upper diagonal, and each row will contain exactly 2 non zero entries $S_s$ and $S_f$ such 
            that $A_{j,Ss} \neq 0, \quad A_{j,Sf} \neq 0, \quad A_{j,Ss}+A_{j,Sf}=1$ and
            $A_{j,Ss}=P(S)_j, A_{j,Sf}=1-P(S)_j$
            
            \item The HMM observation densities $B_j$ are the same as the ABT observation densities.
            
            \item We can thus construct a unique HMM for each ABT.
        \end{enumerate}

%%%%** Section 3.0.2 
\subsection{There is more than one BT for each HMM derived above.}\label{morethan1bt}
\paragraph{Proof:}
        \begin{enumerate}
            \item Element $A_{i,j}$ is defined  above as the probability of a transition from State $i$ to State $j$, however, 
            it does not contain information about whether the transition is associated by the BT
            with \Suc or \Fail of the leaf $i$. 
            
            \item We thus will sometimes augment the HMM $A$ matrix to 
            \[
            A_{i,j,k} = P( i\to j |  O_{i,k}),    k \in \{1,2\}
            \]
            Adding the outcome, $O_{i,k}   \in \{  0,1 \}$, where $O_{i,1} = 1$ if 
            the outcome of leaf $i$ is \Suc, and 
            $O_{i,1} = 0$ if the outcome of leaf $i$ is \Fail. 
            Since  $O_{i,1}, O_{i,2}$ both
            correspond to the \Suc / \Fail outcome of state $i$, and only one is true, 
            we have $O_{i,2} = 1-O_{i,1}$ and either $O_{i,1}$ or $O_{i,2}$ must be 0. 
            
            \item Thus for each row in $A$, containing two non-zero probabilities which add to 1, there are two permutations, S/F and F/S,
            giving more than one BT.
            
            \item When augmented as in Step 2, there is only one BT for each HMM (which follows the constraints  in \ref{UniqueHMM}).
        \end{enumerate}

%%%%** Section 3.0.3 
\subsection{Proposition}\label{sequentialpathway}
\begin{figure}[h]\centering
\begin{subfigure}{0.4\textwidth}
   \includegraphics[width=\linewidth]{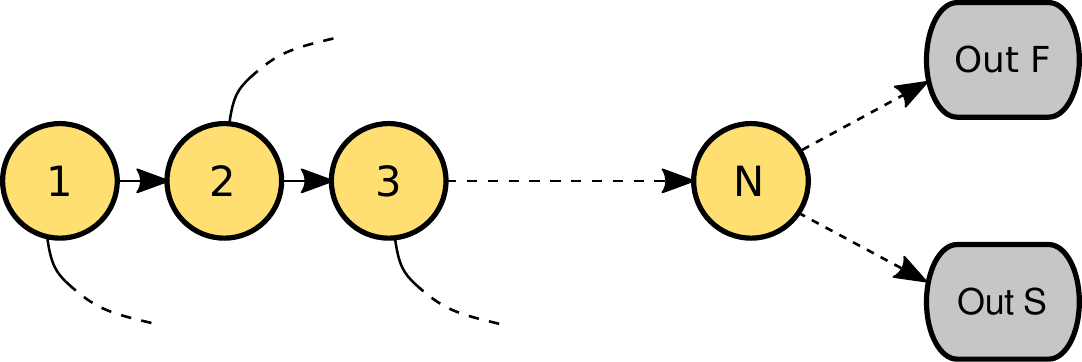}\caption{}\label{BTbasicpathway}
\end{subfigure}
\begin{subfigure}{0.4\textwidth}\centering
   \includegraphics[width=0.4\linewidth]{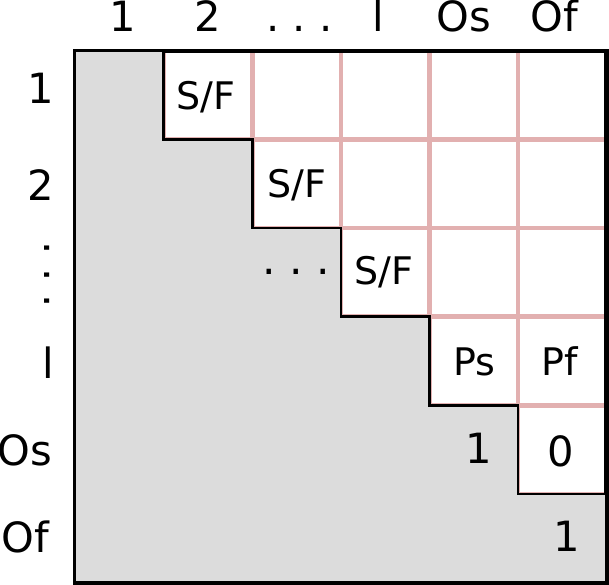}\caption{}\label{AmatrixFig}
\end{subfigure}
{\small \caption{(a) Any BT must have a sequential pathway for a certain set of S/F outcomes.
(b) Constraints on $A$ matrix of HMM derived from BT.}}
\end{figure}

In the state evolution of a BT, there is at least one possible pathway
(determined by S/F outcomes of the leaves) 
in which all $l$ leaves are executed in numerical order (Figure \ref{BTbasicpathway})
such that the BT visits each leaf in turn.

%%%%** Figure 2 

\paragraph{Proof:} 
\begin{enumerate}

    \item Assume all leaves must have at least one input path.  If any leaf has no input path, then it will never
    be executed and should not be in the BT.
    
    \item By convention, BT leaves are numbered from left to right.
    \item Each leaf can belong to a Selector or Sequence parent node.
    \item If leaf $j$ is not the last (rightmost) leaf of its parent then the proposition for 
    leaf $j\to j+1$ is trivial by the definition of Sequence and Selector nodes.
    \item If leaf $j$ is the last (rightmost) leaf of a Sequence or Selector, and control is passed to
    that leaf, then the parent's result is the same as the leaf's result, also by definition of Sequence and Selector nodes.
    
    \item For any Sequence or Selector node which is a parent of leaf $i$, at least one path from that node
 must go to a leaf $j>i$ (unless leaf $i$ is the rightmost node in the entire BT).    
 
    \item If the leaf is not rightmost of the leaves belonging to its parent, then the result is trivial.
 
    \item If the parent IS rightmost, then by induction it will lead either to one of the two Output states (the root node returns and the BT execution is over) 
    or to another higher level node as non-rightmost.
    
    \item Thus there must be a path from node $i$ to $i+1$.
    
\end{enumerate}

%%%%** Section 3.0.4 
\subsection{Implication}\label{HMMConstraints}
Considering these results, the following constraints define the set of
    HMM State-Transition-Matrices ($A$) which correspond to an Augmented BT:
    \begin{enumerate}
        \item $A$ must be upper diagonal (\ref{UniqueHMM} steps 4,5).
        \item $A$ must have exactly 2 non-zero entries per row  (\ref{UniqueHMM} steps 4,5).
        \item $A_{i,i+1} \neq 0$ (\ref{sequentialpathway}).
    \end{enumerate}

% 
%       FUTURE WORK
%

% %%%%** Section 2.0.5 
% \subsubsection{Reverse Mapping: HMM$\to$ BT}
% With the above definition, we can investigate the reverse mapping: Given an $A$ matrix with the above constraints, what
% is the corresponding BT?
% 
% First we consider how many BTs are possible for each $A$ matrix:

%%%%** Section 3.0.5 
\subsection{How many BTs are possible for $l$ BT leaves?}
    \begin{enumerate}
    
        \item Since there is a unique HMM for each BT, we can answer the equivalent question: ``How many HMMs of $l+2$ states meet the constraints of \ref{HMMConstraints}?"
    
        \item Each element just above the diagonal,  $A_{i,i+1}$, must be non-zero because every BT has a sequential pathway. But further, the
        probability in  $A_{i,i+1}$ may represent Success (if leaf $j$ belongs to a Sequence Node) or Failure (if leaf $j$ belongs to a Selector
        Node).  However, the last leaf, leaf $l$, must transition to Os on Success (and to Of on Failure), so there are $2^{l-1}$ permutations of S/F outcomes
        for leaves $i=1..l$ (Fig. \ref{AmatrixFig}).

        \item There must be 2 non-zero entries per row.
        At this step we ignore their Success/Failure status which 
        was taken into account in steps 1 and 2.
        
        \item For row 1, there are $l+2$ columns, but the $A_{1,1}$ element must be zero, and we have already taken into account the 
        $A_{1,2}$ element.   Thus there are  $l+2-2 = l$ available columns, only 1 of which can be non-zero.
        \item For row $i$, there are $l+2-(i+1) = l+1-i$ columns from which to
        independently choose.
        \item Thus there are
        \[
        \Pi (l,l-1, \dots 2, 1) = l!
        \]
        permutations of the second non-zero elements in   rows $1 \dots l$.
        \item Combining, the total number, $m_{BT}$, of BTs for an $l\times l$ $A$ matrix (augmented with the $O_S, O_F$ rows and columns) would
        be
        \[
        m_{BT} = 2^{l-1}l!
        \]

        \end{enumerate}
 
 For 10 leaf nodes there are about 1.86 billion possible behavior trees.

%%%%%%%%%%%%%%%%%%%%%%%%%%%%%%%%%%%%%%%%%%%%%%%%%%%%%%%%%%%%%%%%%%%%%%%%%%%  end of former input

%%%%** Section 3.1
\subsection{Decodability}\label{decodabilitySect}
 As noted above, HMM state decoding can be trivial if the emission probability densities
do not overlap, and impossible if they are identical.  
To achieve a compromise, we defined the  
emission probability over a set of observation symbols 
for state $i$ as a normal distribution with parameters
$\mu_i  = i\times R\sigma, \sigma = 2.0$ and $ \Delta\mu = |\mu_i - \mu_{i+1} |$
where $\Delta\mu$ is a constant and the  ratio $R$,

\beq\label{Reqn}
R = \frac  {\Delta \mu}  {\sigma}
\eeq

is a chosen parameter similar to the Z-score of statistics, which  parameterizes
the degree to which the states are ``hidden'' by the random observations. 
We can term this property the ``decodability" of the HMM.  $R=0$ (no decodability) corresponds to 
completely hidden states and $R=\infty$ corresponds to a trivial state decoding situation (perfect decodability).

%%%%%%%%%%%%%%%%%%%%%%%%%%%%%%%%%%%%%%%%%%%%%%%%%%%%%%%%%%%%%%%%%%%%% 
% New discussion of decodability

A more general notion of the decodability of 
HMM output densities is needed in order to apply the proposed methods to
experimental data which is likely to be non-gaussian.  
For example, if codewords are generated by vector quantization of multidimensional 
sensor signals, they are unlikely to 
be distributed as described above and in particular, unlikely to be confined to a 
limited region (e.g. $\approx \mu_j \pm 4\sigma$ for state $j$) of symbol indeces. 
Thus we will apply Kulback Leibler (KL) and Jensen-Shannon (JS) divergence measures 
to compare the degree to which distributions $B_{i}$
are similar from state-to-state.
Applied to the discrete distributions, we use 
% (the probability of symbol $0<j<NSYMBOLS$ being 
% emitted in state $i$ is $B_{i,j}$) 
the KL divergence for two state observation densities, 
$B_{1,j}$ and $B_{2,j}$ as
\[
D(B_1||B_2) = \sum_{j=1}^{NSYMBOLS} B_{1,j}  \log_2\left(\frac {B_{2,j}} {B_{1,j}}  \right)
\]
Modifying the KL divergence to achieve symmetric and finite properties, we get the JS 
divergence:
\[
JSD(B_1||B_2) = \frac{1}{2}D(B_1||M) + \frac{1}{2}D(B_1||M)  
\]
where $M = \frac{1}{2}(B_1  + B_2)$.  We can also create a measure of the divergence between the 
set of all $N$ observation densities as
\[
JSD(B_1, \dots, B_N) = H \left ( \sum_{i=1}^{N} w_i P_i \right ) - 
\sum_{i=1}^N w_i H(B_i)
\]
where $w_i$ are a set of weights on the densities (uniform, $w_i = 1/N$, in our case) and
$H(P)$ is the Shannon Entropy of distribution $P$.

\begin{figure*}[t]\centering
\begin{subfigure}{0.4\textwidth}
    \includegraphics[width=0.7\linewidth]{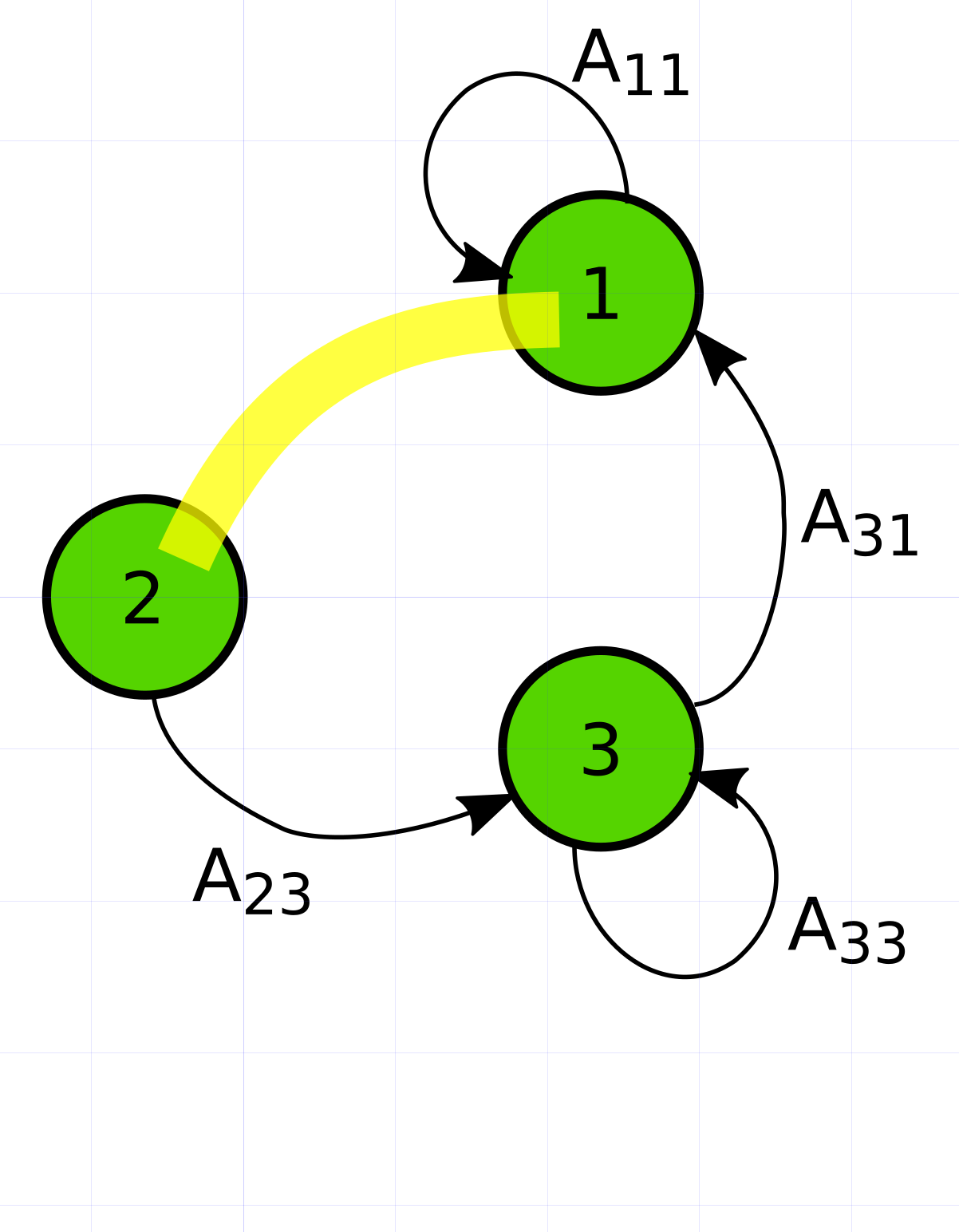}\hspace{0.095\textwidth}\caption{}\label{simplethreestate}
\end{subfigure}
\begin{subfigure}{0.55\textwidth}
    \includegraphics[width=\linewidth]{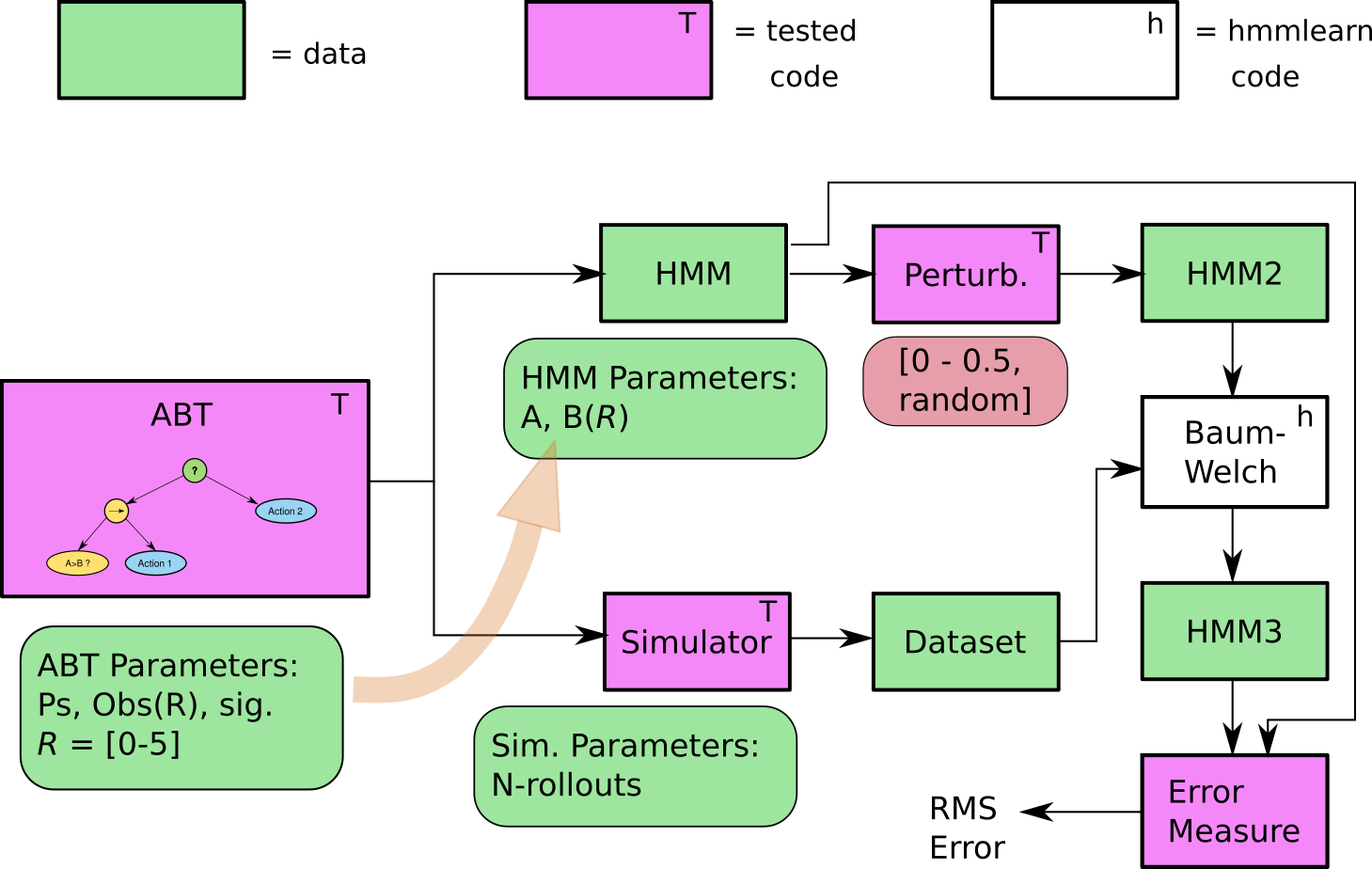}\caption{}\label{BWworkflow}
\end{subfigure}
{\small \caption{(a) Example HMM in which states (separated by yellow arc indicating $D_{12}=0$), and solid
arcs show  state transitions with finite probability.
(b) Workflow for validation of Baum-Welch parameter identification with ABT-derived HMMs using the Baum-Welch algorithm 
of the {\tt hmmlearn.fit()} method. 
The HMM parameters are directly derived from the ABT parameters.
HMM parameters are perturbed from 0-50\% and also randomized.
% Simulation parameter is the number of rollouts simulated.
RMS\_error is the RMS difference of the non-zero elements of two HMM transition matrices.}}
\end{figure*}

Of note, if the KL or JS divergence between two states
is zero, it does not mean the states are completely invisible, depending on the 
transition probabilities.  For example, 
consider a three state model of Figure \ref{simplethreestate} (Left)  in which $JSD_{12}=0$ but $JSD_{13} = JSD_{23} = 0$, and furthermore,  
\[
A = \begin{bmatrix} {}
1&0&0 \\ 0 & 0 & 1\\0.5 & 0 &0.5\\
\end{bmatrix}
\]
Since there is no transition possible between the states with identical observation densities, it is still possible to 
distinguish them. 

%%%%** Figure 3 

%%%%** Section 3.2
\subsection{Behavior Tree Refinements}\label{btrefinement}
The basic combinatorial nodes of the BT are \Seq and \Sel.   
In most of this work we limit consideration of BTs to those composed of leaves and these two 
nodes.    However many BT implementations also use ``Decorator" nodes for additional functions
including

\begin{itemize}  
  \item \Rus nodes 
  \item \Par nodes
\end{itemize}
This section will develop the rules to convert BT sections containing \Rus and \Par nodes into HMMs.  

\begin{figure*}[h]\centering
\includegraphics[width=0.85\textwidth]{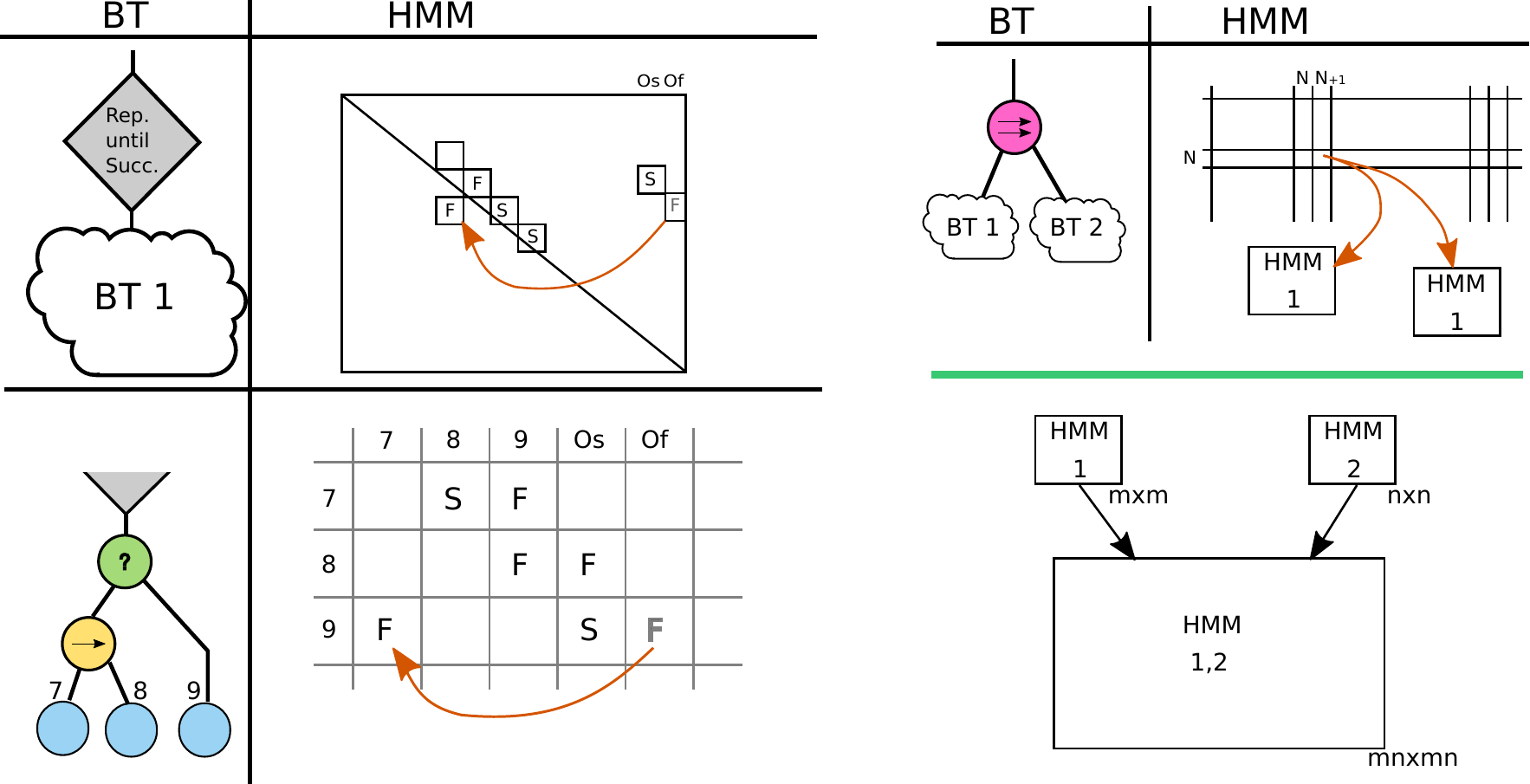}
\caption{Mapping of \Rus \Dec to HMM (Left) results in non-zero terms below main diagonal of HMM state transition matrix, $A$.
Mapping of \Par node to HMM (Right) launches multiple child BTs in parallel, combined 
into a larger HMM submatrix HMM$_{1,2}$.}\label{RUS2hmm}
\end{figure*}

% \paragraph{\Rus \Dec}
\paragraph{Repeat until \Suc Decorator}
This decorator is applied to a child BT and repeatedly restarts that BT until its one child returns \Suc.
It is straightforward to add the \Rus \Dec to a HMM if we relax the constraint from \ref{UniqueHMM}, steps 4,5, that 
the state transition matrix, $A$, be upper diagonal.  A BT which is a child of a \Rus \Dec, BT$_C$, can be inserted 
into the HMM $A$ matrix as follows:
\begin{enumerate}
    \item Assume the first leaf of the child BT$_C$ is numbered leaf $k$ in the main BT (to which it is attached through 
    the \Rus \Dec) and that its subsequent leaves are numbered $k+1, k+2, \dots$. 
    \item Enter the  \Suc and \Fail probabilities for leaves $k+i$ as described in \ref{UniqueHMM}.
    \item Replace any \Fail state transition (e.g. non-zero probability associated with failure)
    in BT$_C$ which would cause BT$_C$ to return \Fail,  with a transition back to state $k$. 
\end{enumerate}
This is illustrated by the example in Figure \ref{RUS2hmm}, Left.   As an example, a three state BT$_C$ where $k=7$ is shown. The 
failure of leaf C, resulting in the failure of BT$_C$ is moved to be a transition back to state 7.
 
%%%%** Figure 4 

% \paragraph{\Par \Dec}
\paragraph{Parallel Decorator}
A \Par \Dec launches multiple child BTs into concurrent execution.   The \Par node 
is configured to return \Suc when a specified fraction of its children return \Suc.
The process to add a \Par \Dec is illustrated conceptually in Figure \ref{RUS2hmm} (Right).  A node corresponding to 
the \Par node is shown at row $N$ with two children, BT1 and BT2.   
The function of the parallel node is to launch multiple 
HMMs corresponding to BT$_C$ trees, BT1 and BT2.
Since the child BTs are executing in parallel, in order to fold them into the main HMM, we must create a set of combined states 
representing the product of HMM1 and HMM2.  To continue this 2-way example, if BT1 has $m$ leaves and BT2 $n$ leaves, the new combined
HMM will have $m\times n $ states. Each new state corresponds to a pair of BT1/BT2 leaves. Each row corresponding to the product-HMM will 
no have 4 non-zero entries corresponding to:

\begin{center}
\begin{tabular}{c|c|c}
 & BT1,$l_i$ & BT2,$l_j$ \\ \hline
1& S & S \\
2& S & F \\
3& F & S \\
4& F & F 
\end{tabular}
\end{center}

Where $l_i$ refers to the leaves of each sub-BT, $1<i<m$, $1<j<n$.
More generally, a parallel node with $K$ children, BT$_k$, each BT having $l_k$ leaves, will generate a square submatrix in the overall BT's $A$ matrix 
having 
\[
\displaystyle\prod_{k=1}^K l_k
\]
states (rows and columns).  

This significant expansion of the number of states in the HMM, as well as the increase from 2 to 4 non-zero entries in rows, 
may limit the practical usefulness of HMMs derived from BTs which make use of parallel nodes.

%%%%** Section 4 
\section{Computational Experiments}\label{methodssection}

In this section we    evaluate the ability of HMM-derived algorithms to work 
with ABTs.   Ultimately, as with HMMs, much depends on the values of HMM/ABT 
parameters, especially the degree to which the states are ``decodable" (Section \ref{decodabilitySect}).
% For example, as in Section \ref{decodabilitySect}, 
% if the emission density is identical for each state, 
% then recovering the state sequence is fundamentally impossible.  On the other 
% hand, if the emission densities of the states do not overlap at all, then the 
% recovery of state sequence is trivial.   The objective of these experiments then 
% is to characterize performance of the three classic HMM algorithms on  ABTs with 
% varying degrees of this obscurity.  
To parameterize the feasibility of these computations,
we take two approaches.   

First,  we perturb the HMM parameters prior to their
use with the
HMM-based algorithms.  For example, we test the ability of the Viterbi decoder to 
identify the sequence of ABT/HMM states with a model that is perturbed  from the model which
generated the ABT rollouts and observations.
The perturbation was applied to each row of the HMM transition matrix as follows:
for a perturbation $0 <\tilde{p} < 0.5$, 
let $p_1$ be the first (lowest index) non-zero element of each row, and $p_2 = 1-p_1$ is the second
        non-zero element.  Then the two non-zero elements are modified as 
        $p_1 \gets \max(0.95, (1 + \tilde{p}\times r) p_1)$
        and
        $ p_2 = 1-p_1 $,  $ r \in \{-1, +1\}$ is a random sign variable drawn
        with 0.5 probability on each sign. 
Because it is frequently done in HMM research, for comparison we created another 
HMM transition matrix
by initializing all HMM elements to random values [0.0-1.0) (subject to each row totaling 1.0).
Second we study performance under various levels of ``decodability" of the HMM states, 
by generating data sets with values of $R$ from the set $\{0,0.25, 1.0, 2.5, 5.0\}$.
Simulations were generated from  6-state and 16-state BTs to explore the effect of model size. 

% %%%%** Section 4.1
% \subsection{Applying well known  HMM algorithms to ABTs}
% 
% HMMs have been used for decades to analyze time series data.
% In this section we evaluate the ability of the three
% ``classic" HMM algorithms of Section \ref{ModelReview}\cite{rabiner1989tutorial}
% to perform their functions on data arising from simulations of ABTs. The goal 
% is to evaluate the effectiveness of modeling and 
% tracking the evolution of a human or a system,
% described by a BT and measured or observed through a stochastic process. 
% 

% \paragraph{Simulation Setup}
\label{SimSetupSection}

For the two ABTs, we generated simulated traces of 15,000 state (leaf) sequences
and their associated random emissions by running ABT Monte-Carlo simulations.  
Results did not 
change when more than 15,000 sequences were generated. 
State evolution of the ABTs was determined by the success/fail probability of each leaf,
$ps_i$ (Section \ref{ABTdefs}), and emissions determined by the observation
density associated with each leaf $B_{ij}$. 
We implemented the simulation study in Python 2.7 using a modified version of the {\tt b3} Behavior Tree library\cite{marzinotto2014towards,b3library}. We added a new leaf class to 
{\tt b3}, {\tt abt\_leaf(b3.Action)} which inherits the {\tt Action} class (BT leaf) and adds 
data to represent the probability of \Suc and \Fail, as well as a probability of 
discrete observations/emissions from the leaf.

%%%%** Section 5 
\section{Results}

\subsection{Relationship to idealized emission probability densities}
We computed the KLD \& JSD divergence measures  for the $B_{ij}$
and compared  them with several values of $R$, eqn(\ref{Reqn})  ( Table \ref{TableRProj}).
The maximum theoretical value of $JSD_{ALL}$ is $\log_2(N)$ where $N$ is the number of 
HMM states ($l+2$) in our application.   For the two cases in  Table \ref{TableRProj},
these values are $\log_2(6) = 2.585$, and $\log_2(16) = 4.0$.
Thus when $R=5$ the observation distributions are essentially  distinct.

%%%%** Table 1 
\begin{table*}\centering
\caption{Measures of ``decodability". Relationship between  ratio $R$ (\ref{Reqn}), and 
$KLD$, $JSD$ divergences between two consecutive state distributions, 
and $JSD_{ALL}$ between all pairs of states. 
Measures are evaluated for both  6-state and 16-state models.}\label{TableRProj}
% \begin{tabular}{|l|l|l|l|l||l|l|l|l|l|}\hline
\begin{tabular}{llllllllll}\hline
%%%
 Ratio& N & $KLD$ & $JSD$ & $JSD_{ALL}$ &        N  & $KLD$ & $JSD$ & $JSD_{ALL}$\\ \hline
 0.00 & 6 & 0.00 & 0.00 & 0.00 &                16 &  0.00 & 0.00 & 0.00 \\ \hline
 0.25 & 6 & 0.18 & 0.04 & 0.12 &                16 &  1.62 & 0.32 & 0.60 \\ \hline
 1.00 & 6 & 2.89 & 0.49 & 0.95 &                16 &  22.87 & 0.99 & 2.11 \\ \hline
 2.50 & 6 & 17.47 & 0.98 & 1.98 &               16 &  26.85 & 1.00 & 3.32 \\ \hline
 5.00 & 6 & 26.85 & 1.00 & 2.54 &               16 &  26.85 & 1.00 & 3.95 \\ \hline
%%%
\end{tabular}
\end{table*}

%%%%** Section 5.1
\subsection{Forward Algorithm}
 
%%%%** Figure 8 
\begin{figure}[h]\centering
\includegraphics[width=0.5\textwidth]{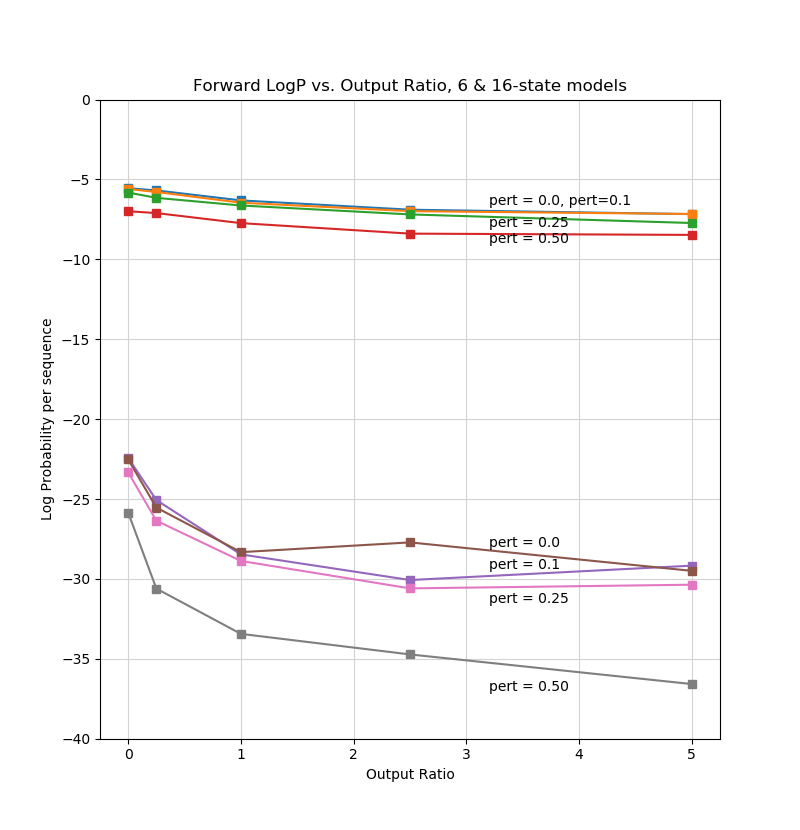}
\caption{HMM Forward Algorithm: Log Probability with respect to simulated data vs.
Output Ratio and perturbation. 
{\bf Upper Traces:}  6-State Model, 
{\bf Lower Traces:}  16-State Model.}\label{res_fwd_alg}
\end{figure}

The forward algorithm was applied to 20,000 runouts of observation symbol data
generated by several 6-state and 16-state ABT-like HMMs in which the output 
Observation Ratio, $R$, was varied between 0 to 5.0 to represent the completely hidden
state case ($R=0$) and the transparent state ($R=5$) cases. 

The models used to
generate the runout data were perturbed by [0, 0.1, 0.25, and 0.5] ratios (Section \ref{methodssection}).
For each combination of output ratio and perturbation ratio, we computed the log probability (logP) of
observing each sequence in the simulated data set. Log probability was divided by number of observation runouts to get an average log probability per sequence (Figure \ref{res_fwd_alg}).
For the 6-state ABT-like model, logP per sequence was much higher than for the 16-state model. 
As expected, logP declined as the perturbation (indicating distance between the data generating and the 
data evaluation models) increased.  logP declined as output ratio, $R$, increased (approx -15\% for 
6-state model and -30\% for 16-state model).   
This decline may seem counter intuitive, but perhaps is explained by the state transition sequence having 
less flexibility for the forward algorithm to fit.

\subsection{Viterbi State Decoding}

Simulated state evolution data, generated by a reference ABT model,
was read by the {\tt hmmlearn} Viterbi decoder.  The state sequence produced was then
compared with the true state sequence  using
the string edit distance (Levenshtein distance \cite{levenshtein1966}) between the 
estimated and true state sequences divided by the number of symbols.  
% The final score from this test is the average string edit distance of the state sequences. 
The expected result is that Viterbi decoding by the  HMM with zero perturbation of parameters will give the 
lowest average string edit distance. The simulation data was also input to 
the Baum-Welch algorithm to assess parameter identification performance when initialized 
with a perturbed version of the reference model.
 
\newcommand{\sfw}{0.4\textwidth}
\newcommand{\bfw}{\begin{figure*}}
\newcommand{\efw}{\end{figure*}}

%%%%** Figure 5 
\bfw[h]\centering
\begin{minipage}[c]{\textwidth}\centering
\begin{subfigure}{\sfw}\centering
    \includegraphics[width=\linewidth]{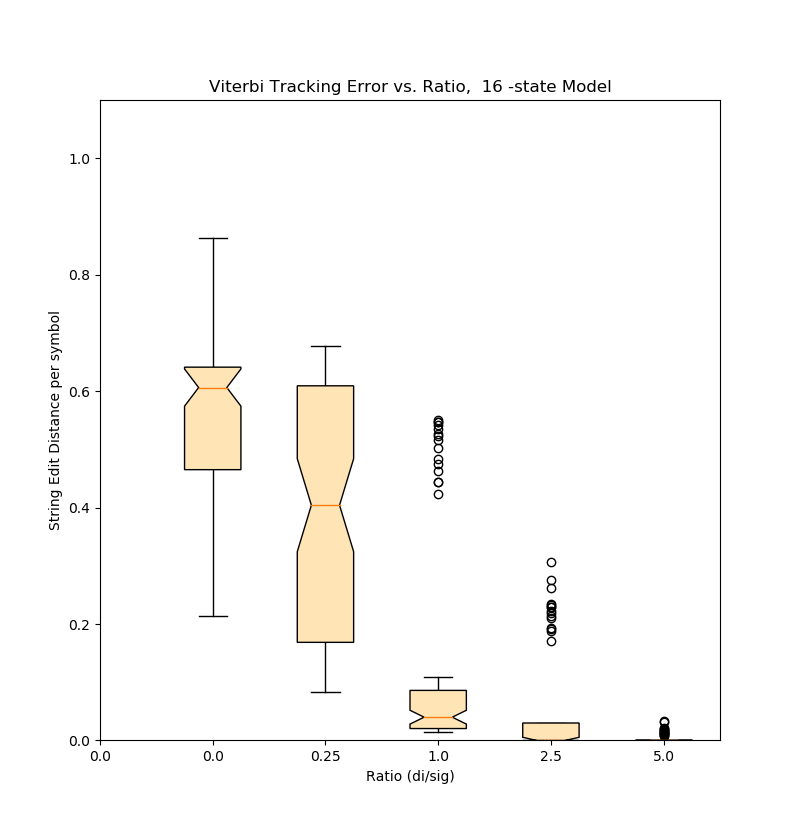}\caption{}
\end{subfigure}
\begin{subfigure}{\sfw}\centering
    \includegraphics[width=\linewidth]{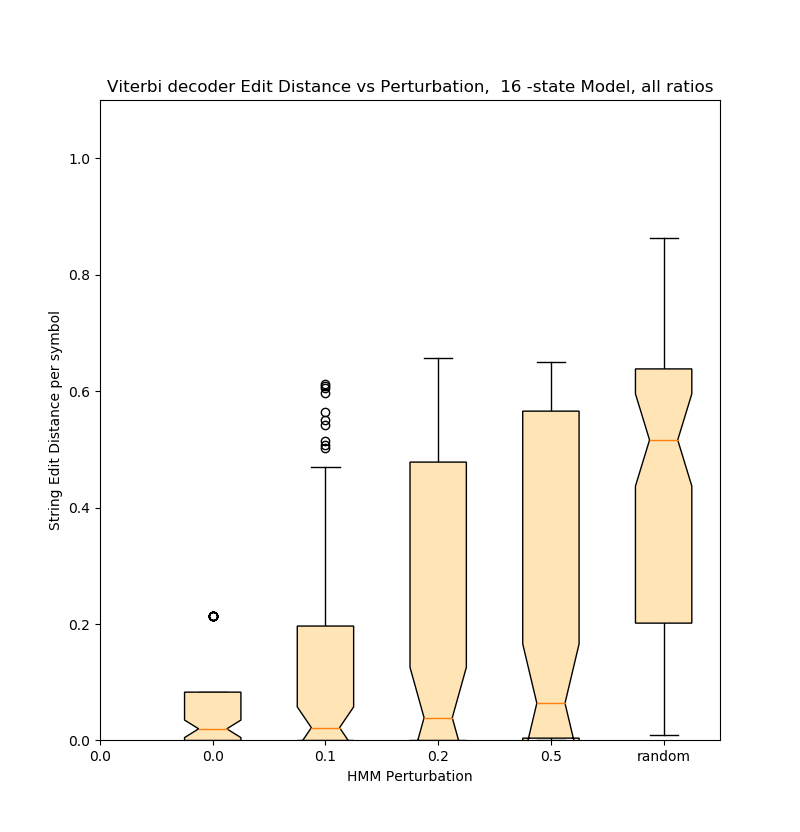}\caption{}
    \end{subfigure}
\newline
\begin{subfigure}{\sfw}\centering
    \includegraphics[width=\linewidth]{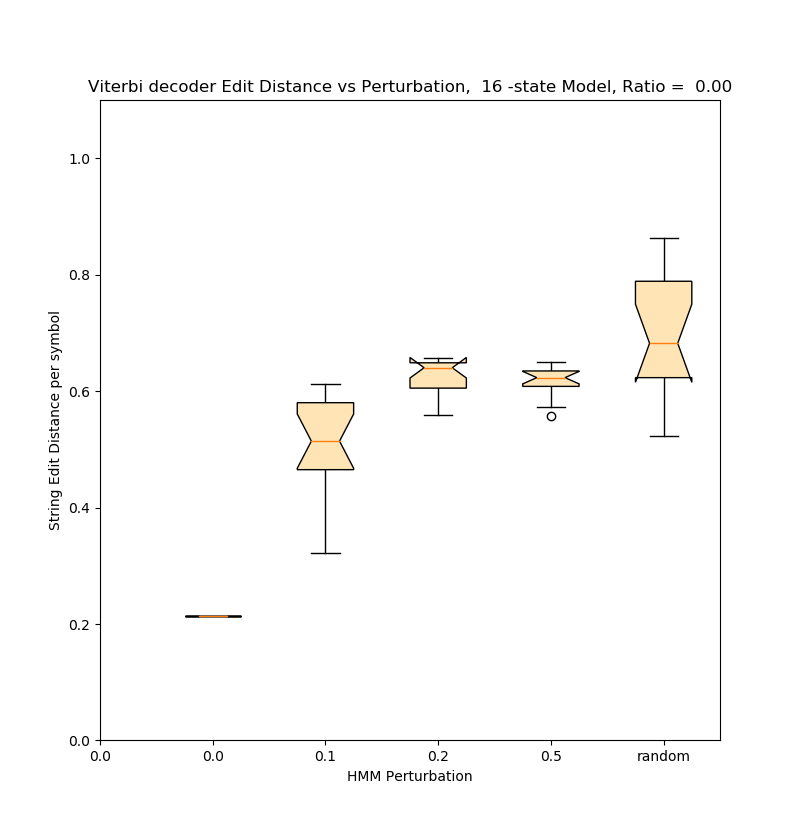}\caption{}
\end{subfigure}
\begin{subfigure}{\sfw}\centering
    \includegraphics[width=\linewidth]{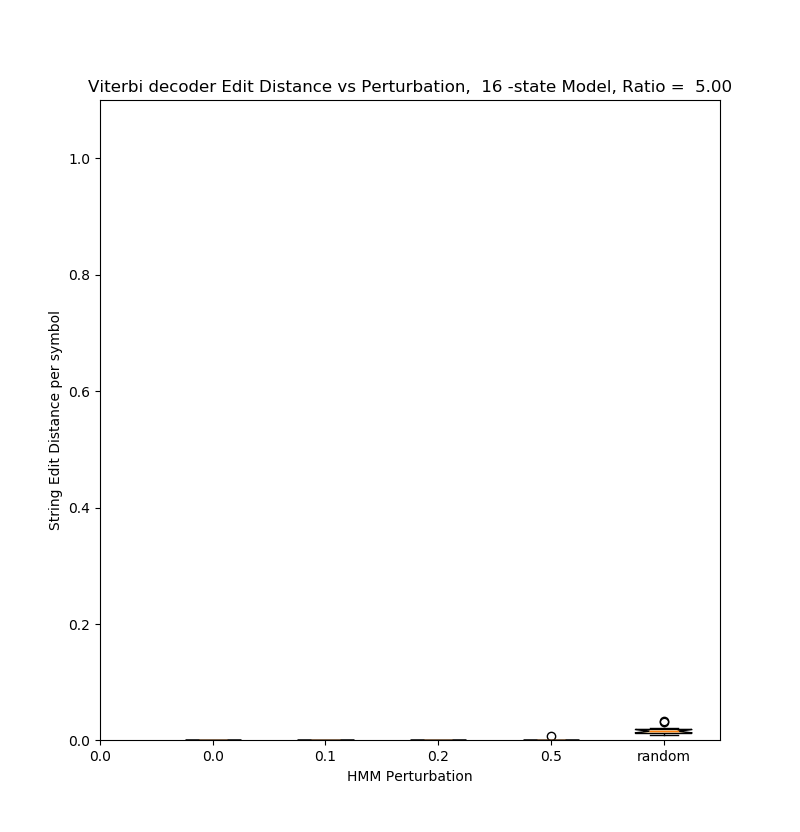}\caption{}
\end{subfigure}
\end{minipage}
\caption{Viterbi Algorithm state tracking error (string edit distance per symbol, SED) of 16 state model.
{\bf Upper Left:}  SED vs Ratio, $R$, for all model perturbations.
{\bf Lower Left:} ``hard" models ($R=0$, all states have same observation density).
{\bf Upper Right:} SED vs Model perturbation, all ratios.
{\bf Lower Right:} ``easy" models ($R=5$, symbols are 5 standard deviations apart).}\label{res_SED_16state}
\efw
% 
% 
% %%%%** Figure 5 
% \begin{figure*}
% \includegraphics[width=0.5\textwidth]{Result_Figures/res_Vit_vs_R_16_allratios.png}
% \includegraphics[width=0.5\textwidth]{Result_Figures/res_Vit_vs_P_16_allratios.png}
% \includegraphics[width=0.5\textwidth]{Result_Figures/res_Vit_vs_P_16_Ratio-0p00.png}
% \includegraphics[width=0.5\textwidth]{Result_Figures/res_Vit_vs_P_16_Ratio-5p00.png}
% \caption{Viterbi Algorithm state tracking error (string edit distance per symbol, SED) of 16 state model.
% {\bf Upper Left:}  SED vs Ratio, $R$, for all model perturbations.
% {\bf Lower Left:} ``hard" models ($R=0$, all states have same observation density).
% {\bf Upper Right:} SED vs Model perturbation, all ratios.
% {\bf Lower Right:} ``easy" models ($R=5$, symbols are 5 standard deviations apart).}\label{res_SED_16state}
% \end{figure*}

\paragraph{16-State Model}

%%%%** Figure 6 
\bfw[h]\centering
\begin{minipage}[c]{\textwidth}\centering
\begin{subfigure}{\sfw}
    \includegraphics[width=\linewidth]{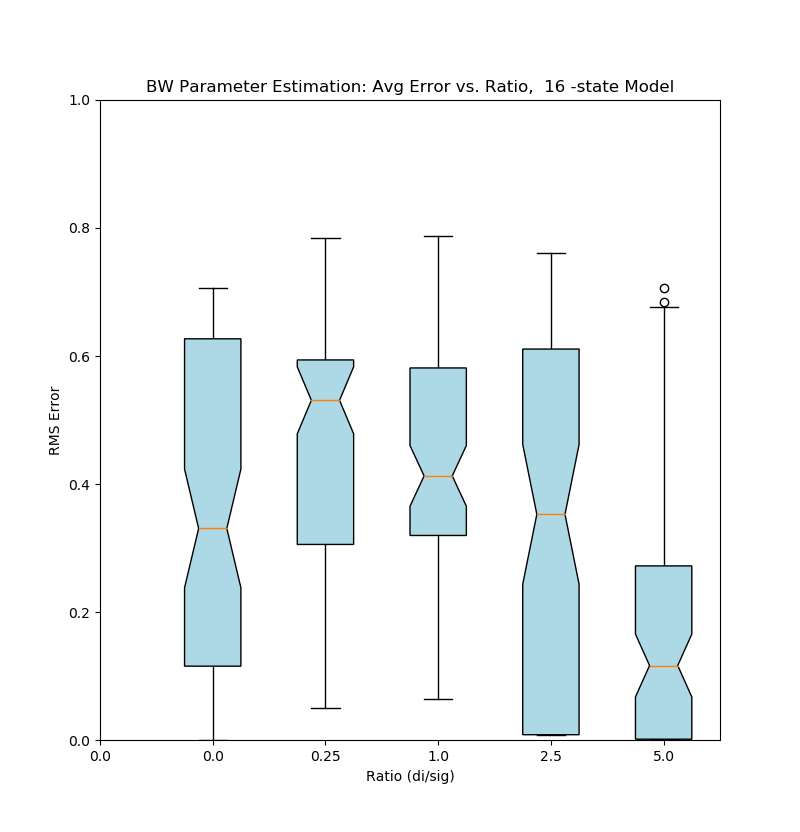}\caption{}
\end{subfigure}
\begin{subfigure}{\sfw}
    \includegraphics[width=\linewidth]{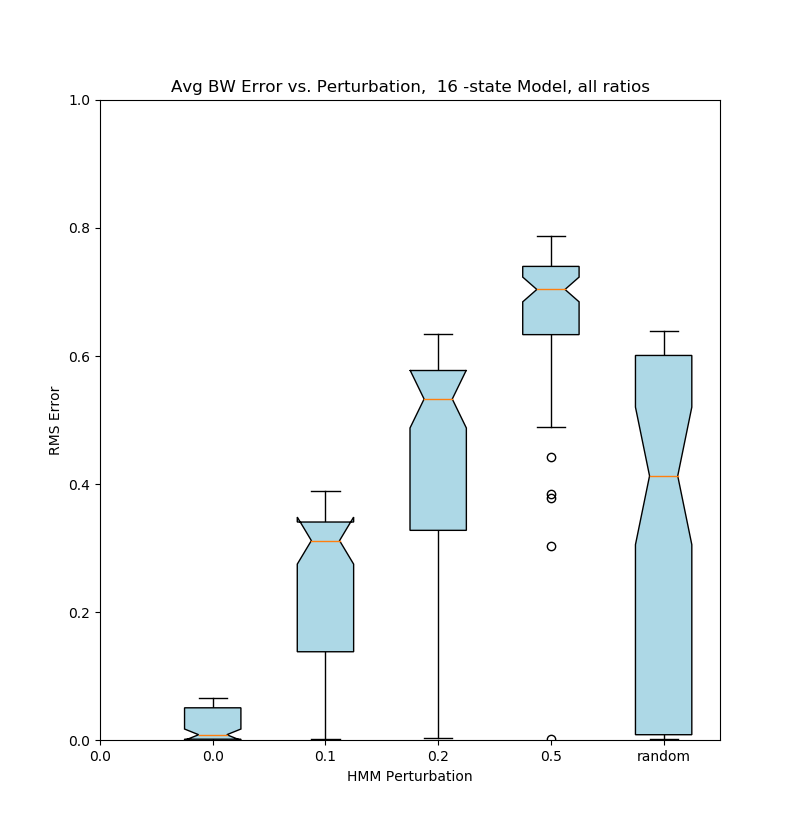}\caption{}
    \end{subfigure}
\newline
\begin{subfigure}{\sfw}
    \includegraphics[width=\linewidth]{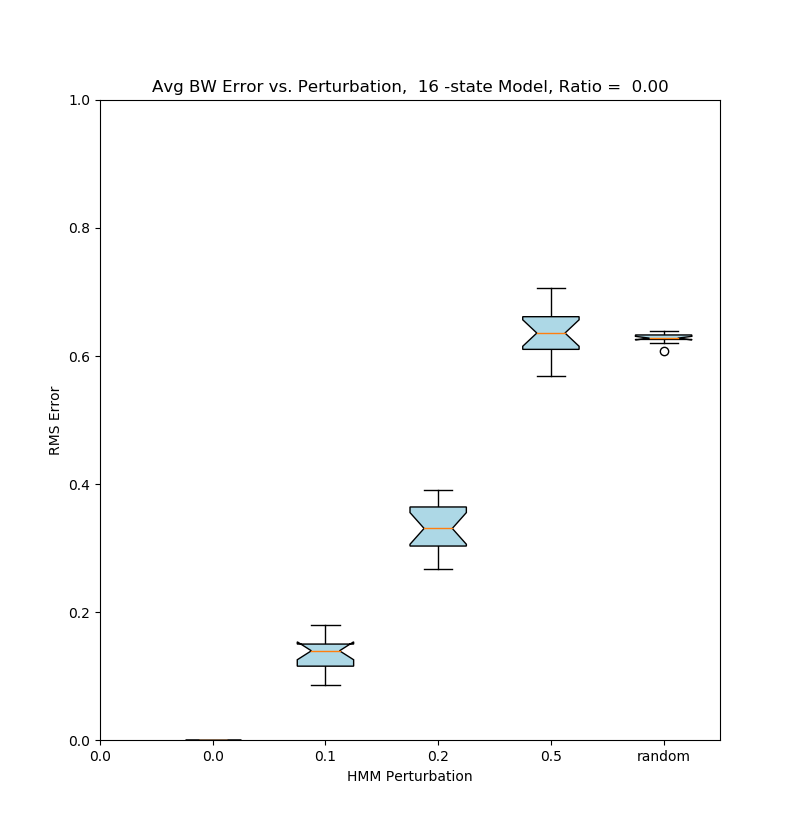}\caption{}
\end{subfigure}
\begin{subfigure}{\sfw}
    \includegraphics[width=\linewidth]{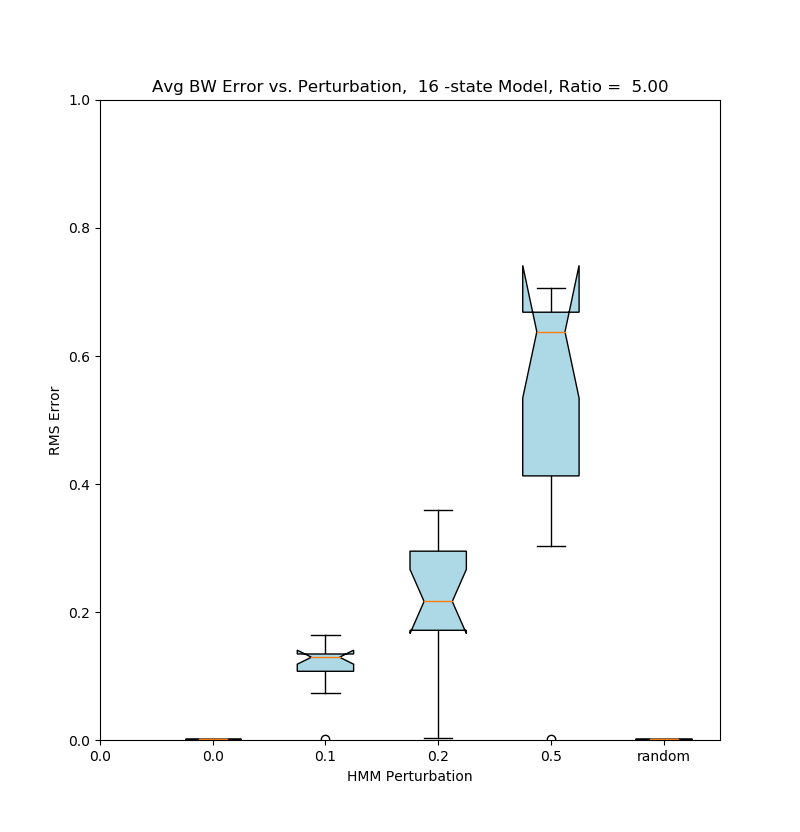}\caption{}
\end{subfigure}
\end{minipage}
\caption{Performance of the Baum-Welch Parameter Identification algorithm on a 16 state model against ABT derived sequence data set and associated HMM.   
{\bf Upper Left:} BW errors vs Ratio, $R$, for all model perturbations.
{\bf Lower Left:} ``hard" models ($R=0$, all states have same observation density).
{\bf Upper Right:} Error vs Model perturbation, all ratios.
{\bf Lower Right:} ``easy" models ($R=5$, symbols are 5 standard deviations apart).}\label{16StateBWResults}
\efw 

% 
% %%%%** Figure 4 
% \begin{figure}
% \includegraphics[width=\textwidth]{nips_perf2.png}
% \caption{Viterbi Algorithm state tracking error (string edit distance per symbol, SED) of 16 state model.
% {\bf Upper Left Yellow (ULY):}  SED vs Ratio, $R$, for all model perturbations.
% {\bf LLY:} difficult models ($R=0$, all states have same observation density).
% {\bf URY:} SED vs model perturbation, all ratios.
% {\bf LRY:} easy models ($R=5$, symbols are 5 standard deviations apart).
% (4 Right/blue plots)  
% {\bf ULB:} BW errors vs Ratio, $R$, for all model perturbations.
% {\bf LLB:} difficult models ($R=0$, all states have same observation density).
% {\bf URB:} Error vs Model perturbation, all ratios.
% {\bf LRB:} easy models ($R=5$, symbols are 5 standard deviations apart).}\label{res_SED_16state}\label{16StateBWResults}
% \end{figure}

Referring to the yellow box plots of Figure \ref{res_SED_16state}, state tracking performance
measured by average Levenshtein distance, improves as the observation ratio, $R$, increases
(essentially perfect, as expected, for $R=5$), and degrades as the model is perturbed away from the 
reference model. 

Parameter identification using the Baum-Welch algorithm (blue box plots of Figure \ref{res_SED_16state})
was  a strong function of the perturbation away from the reference model for all values of $R$.
As expected, the BW algorithm did not change the model if it was initialized with the exact reference value (perturbation = 0). A random initialization gave highly variable results but it should be noted
that the RMS error measure only compared $A_{i,j}$ elements which are non-zero in the reference model.
Thus the adapted randomized $A_{i,j}$ matrix could still be expected to be very different. 

% 
% 
% Figure \ref{res_SED_16state} (Yellow (Y), Upper Left) shows   SED results
% for the 5 values of observation Ratio, $R$ in the 16-state model.
% All HMM perturbation values are included in this data set.  
% Figure \ref{res_SED_16state} (Y, Upper Right) shows average SED for 4 values of HMM perturbation, 
% and also for a  random HMM $A$ matrix for all Ratios in the 16-state model. 
% There is a notable trend of worse state tracking results for higher model perturbations 
% and especially for the random HMM state-transition matrix. 
% Looking specifically at the lowest model state transparency ($R=0$), (Figure \ref{res_SED_16state}, Y Lower Left)
% performance is only good when the model is exact (HMM perturbation = 0).   In contrast, when 
% model transparency is highest ($R=5.0$), state tracking is extremely accurate for all model perturbations (as
% expected). 
%  
% 
% Breaking out the Viterbi tracking results according to the observation ratio, $R$, as expected
% the state tracking performance is poor for $R=0$ (Figure \ref{res_SED_16state} (Y, Lower Left))
% and excellent when $R=5.0$ (Figure \ref{res_SED_16state} (Y, Lower Right)).
% % Recall that states are completely hidden for $R=0$ and have 
% % essentially non-overlapping observations for $R=5$.
% 
% 
\paragraph{6 State Model}   Performance of the Viterbi state tracking algorithm on the 6-state model 
and its dependence on the Ratio and perturbation parameters 
was qualitatively similar to that of the 16-state model, but errors were lower overall
and relatively independent of perturbation.

\begin{figure}[h]
\includegraphics[width=0.5\textwidth]{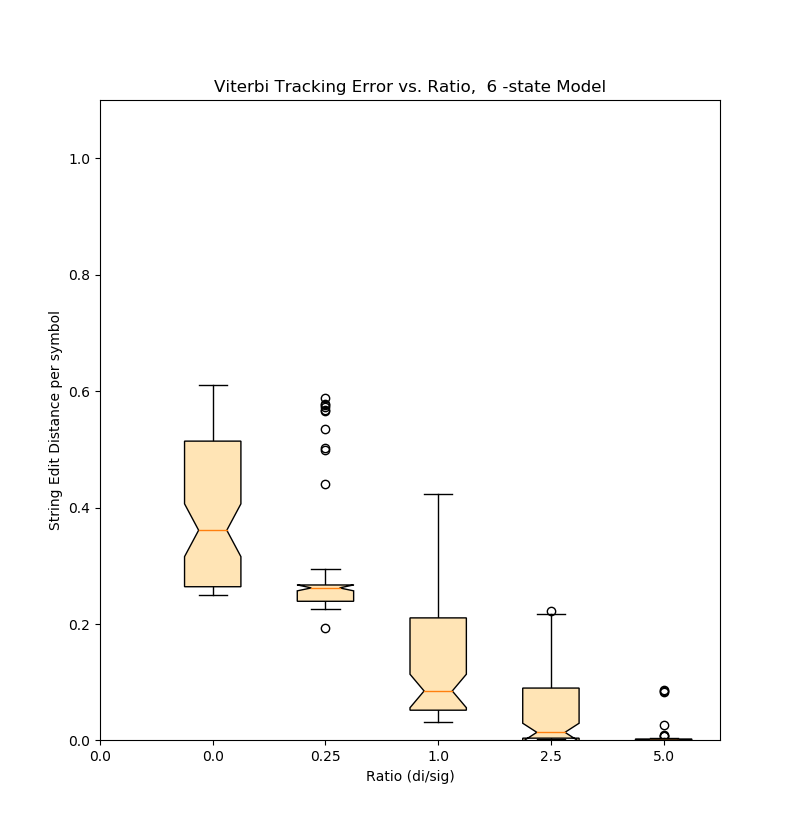}
\includegraphics[width=0.5\textwidth]{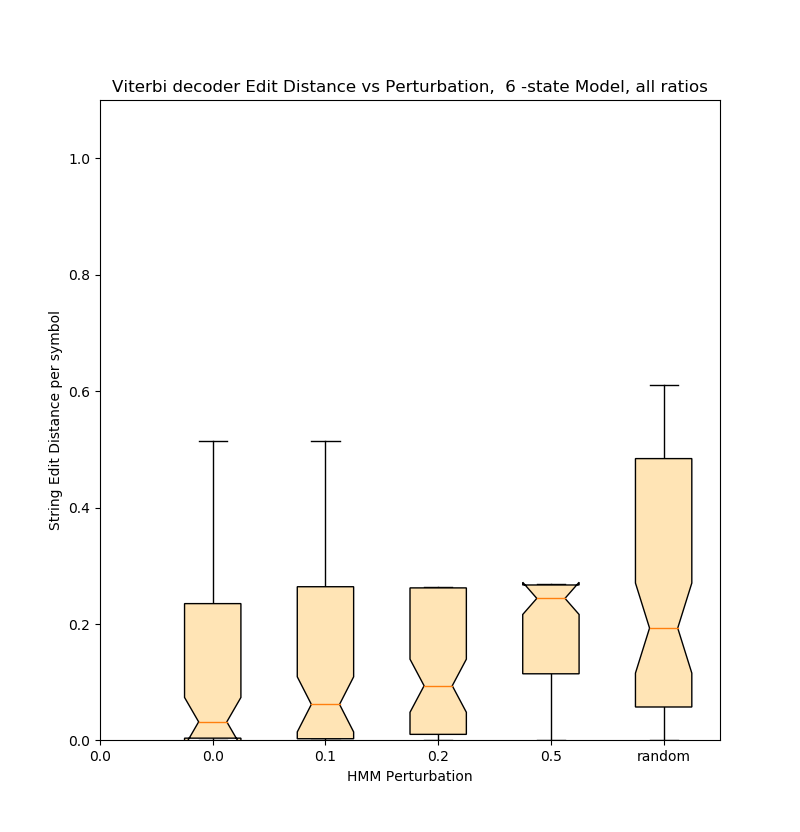}
\includegraphics[width=0.5\textwidth]{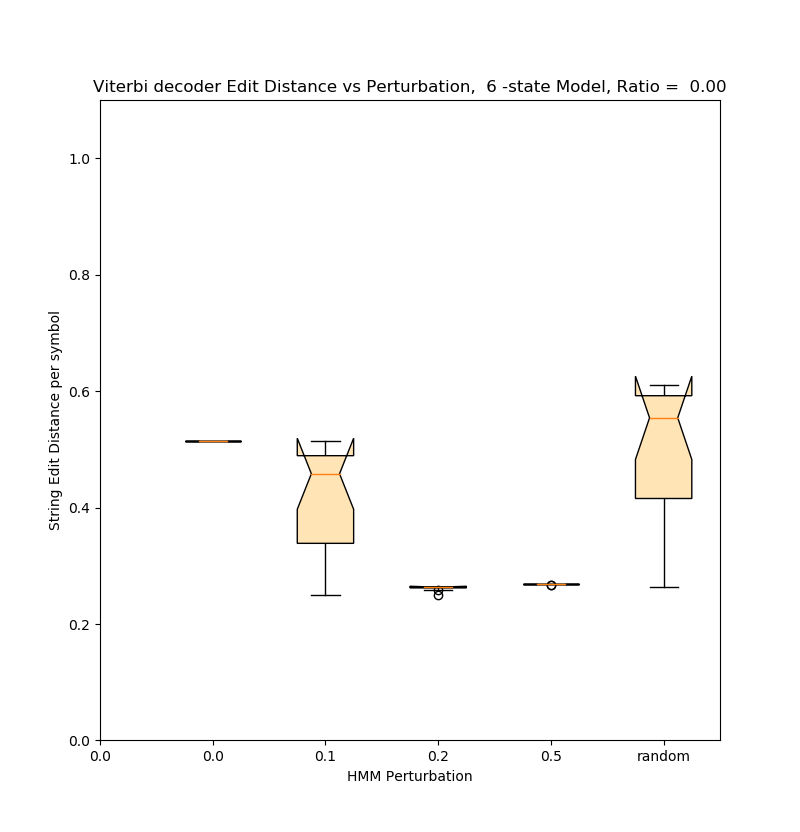}
\includegraphics[width=0.5\textwidth]{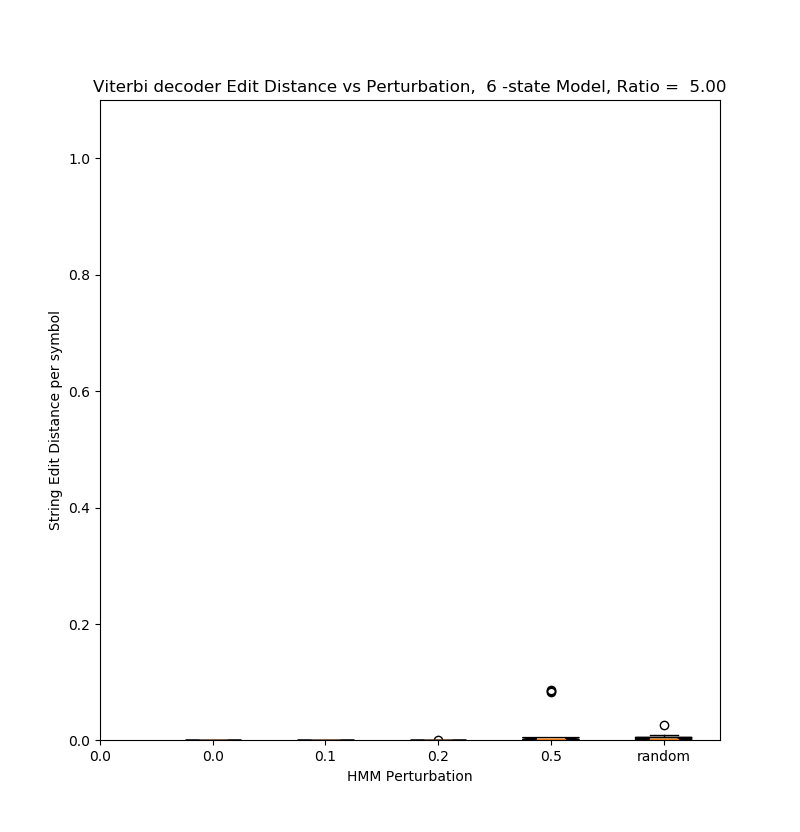}
\caption{Viterbi Algorithm state tracking error (SED) vs. Ratio (Left) and vs. Perturbation (Center).  
Graph details are same as
Figure \ref{16StateBWResults} The 6-state model has 12/36 non-zero elements.}\label{res_SED_6state}
\end{figure}
 
%   

%%%%** Section 5.2
\subsection{Baum-Welch Parameter Identification}
 
Finally, the {\tt hmmlearn} Baum-Welch algorithm  
({\tt HMM.fit()}) was applied to the simulation-generated evolution file to test the 
ability to identify HMM parameters from data.   In some of our experiments, we initialized
the HMMs to random values because many previous studies have used randomly initialized parameters.   

Baum Welch parameter estimation was evaluated on 300 randomly generated ABT-derived HMMs. 
We evaluated BW performance primarily in terms of 1) RMS error between the HMM parameters used
to generate data and the final HMM parameters identified by BW and
2) The log probability (``LogP") of the model after BW adaptation as a function of parameters such as model size and the amount of perturbation of the A-matrix. 

\begin{figure}
\includegraphics[width=0.5\textwidth]{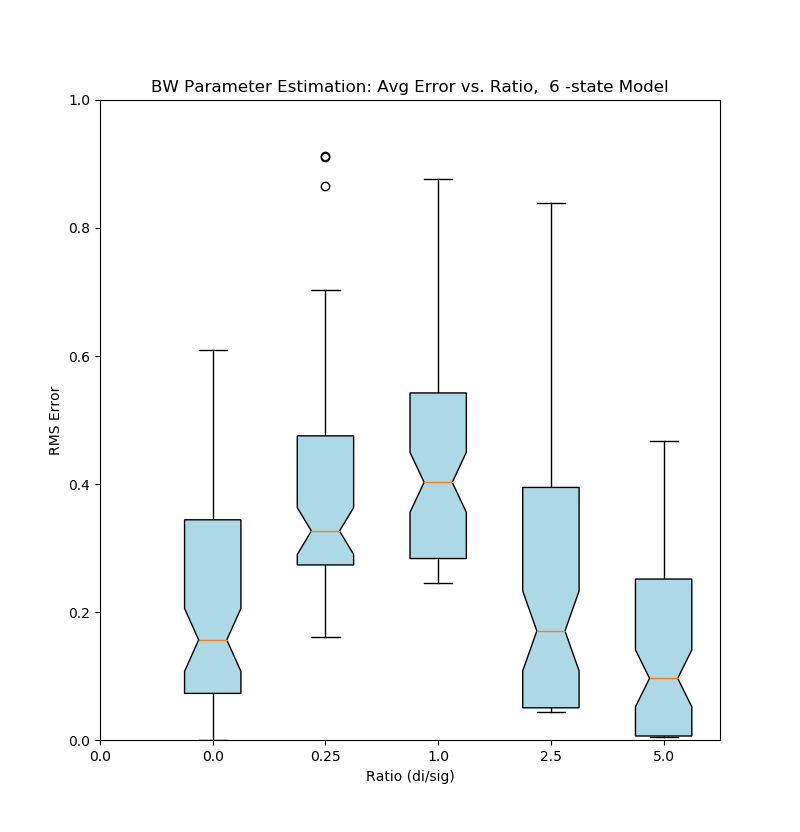}
\includegraphics[width=0.5\textwidth]{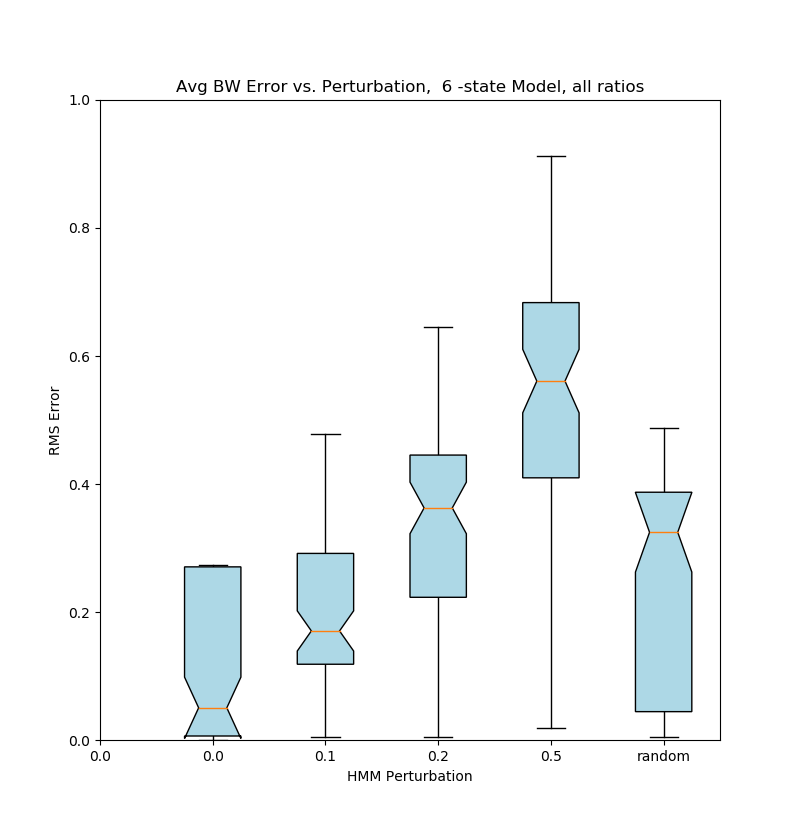}
\includegraphics[width=0.5\textwidth]{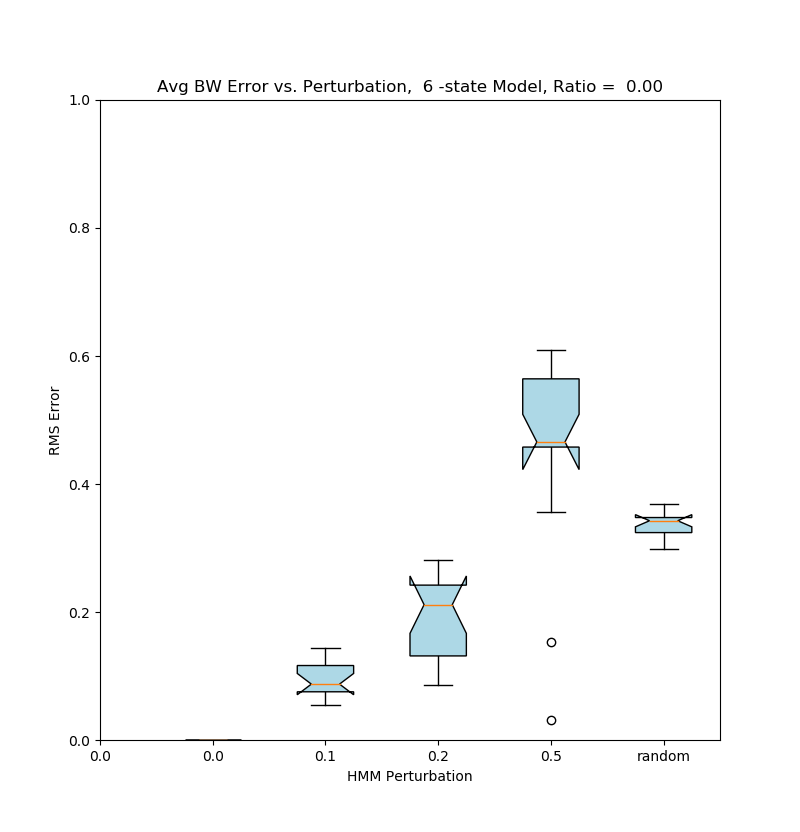}
\includegraphics[width=0.5\textwidth]{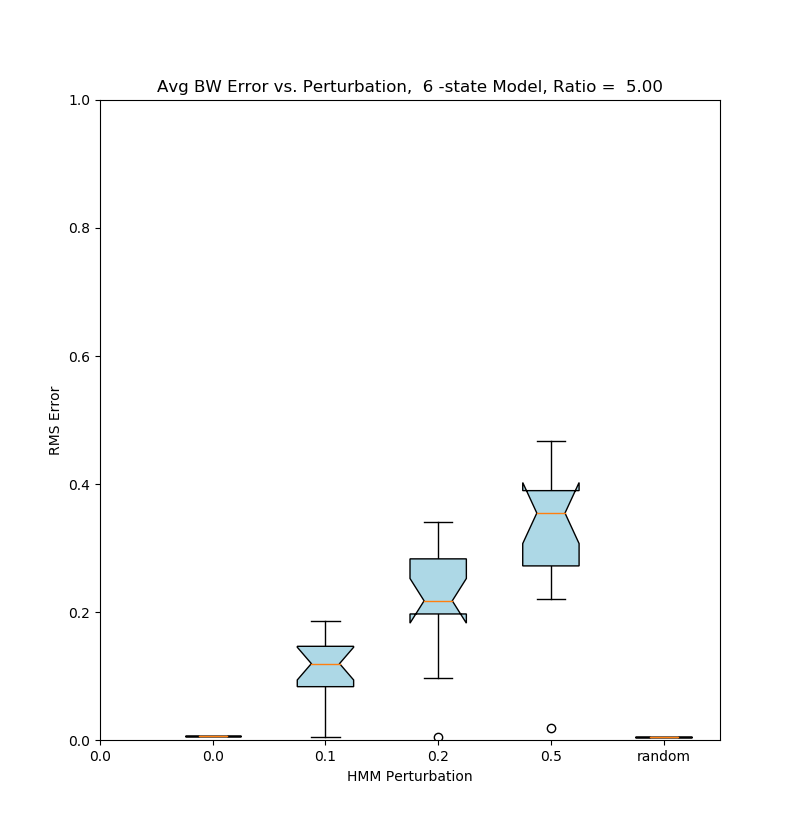}
\caption{Performance of BW algorithm for 6-state ABT derived data set. Graph details are same as
Figure \ref{16StateBWResults}. The 6-state model has 12/36 non-zero elements. }\label{6StateValResults}
\end{figure}

\paragraph{6-State Model}  
For the 6-state model (not shown), 
BW identification worked much the same way although improvement in parameter identification was greater
for 2.5 and 5.0 ratios with the smaller model. 
Average RMS\_error vs. perturbation was better for all ratios with the 6-state model compared to the 16 state model.
For the 6-state model with obscured states ($R=0$) parameter estimation improved with respect to the 16-state
model, especially for the random initialization. 
For the 16-state model with transparent states ($R=5$) parameter estimation was better. 

In both small and large models, parameter estimation error was minimal with transparent states. 
This is expected because the state
transitions are unambiguous for those models. 
\section{Discussion and Future Work}
 
\paragraph{Relation between BT and HMM}   In this paper
we have elucidated a  relationship (to our knowledge new) between Behavior Trees and HMMs such that 
if the BT is augmented with probabilistic information,  it then corresponds to a unique HMM. 
HMMs have been extremely successful in intelligent processing of time series data in fields such as
speech, robotics and surgical manipulation.   The anticipated applications for this result include
processing and analysis of time series data, corrupted by noise or stochasticity, but arising from a 
process which can be described by a BT.  Such processes include both autonomous agents and humans 
performing tasks with real-world interaction. 

In such an application, this method could be applied as follows:
1) model the process with a BT according to expert knowledge
2) collect experimental data and manually label it with BT state (i.e label each observation symbol with the 
corresponding process step). Such labeling can 
be done by process experts or crowd sourced workers  with or without information not present in the 
observation such as video or direct observation.
3) compute ABT statistics from the labeled data including $A_{i,j}$ and $B_{ij}$.
4) evaluate decoding/identification feasibility with divergence measures.
5) generate a HMM from the ABT
6) conduct the required data analysis using existing HMM algorithms. 

In this paper, the emission probability density of each leaf was independent of the \Suc or \Fail outcome.
It may be that a modification in which the emission density for each state is conditioned on the
two outcomes of each state would be more realistic in some applications.

\paragraph{Divergence measures.}  We computed divergence measures for sets of 
artificial $B_{ij}$ distributions having
different degrees of overlaps: an admittedly simplistic discrete Gaussian probability distribution.
In an application, the divergence measures will be used to predict the effectiveness
of using the methods of this paper on a new data set.
Using the labeled data set (steps 2,3 above), we can compute the preferred divergence measure from the estimated $B_{ij}$.  The empirical divergence measure thus obtained, along with Table 1, 
can relate a new data set to the experimental results in (Figure \ref{res_SED_16state}).

\paragraph{Viterbi state sequence decoding} The Viterbi algorithm 
worked well for most conditions (Figure \ref{res_SED_16state} yellow box plots), 
especially when Ratio was greater than 1.0 ($JSD \approx 1.0$). With high divergence (``decodability"),
decoding was perfect even with substantial difference between generative and decoding model 
parameters.

\paragraph{Baum-Welch parameter identification} Performance of BW parameter identification
was more mixed (Figure \ref{16StateBWResults} blue  box plots).  Results show that error in the HMM parameters was proportional to the 
difference between data generation and decoding models.  For comparison, RMS error due to the 
model perturbation was about 0.1 compared to the 0.2-0.6 seen after BW convergence. This was true even
for models with high divergence.

\paragraph{Implications for future work}
Linking BTs to HMMs allows classic HMM algorithsm to be used as probabilitic tools for systems which
can be modeled by BTs.   Since HMMs in turn are linked to Dynamic Bayesian Networks 
(DBNs), an even wider set of tools are potentially  available. In particular, this may improve model
identification 
as some DBN techniques show better performance than Baum-Welch under certain conditions \cite{ghahramani2001introduction,hamid2003argmode}.

\subsubsection*{Acknowledgments}

The author would like to thank Shao-An (Sean) Yin for suggesting divergence measures and 
acknowlege support from 
NSF Grant IIS-1637444, and the University of Washington / Tsinghua University Global Innovation Exchange, GIX
  
\bibliography{Master}
\end{document}